\documentclass{article}
\usepackage{ijcai26}

\usepackage{times}
\usepackage{soul}
\usepackage{url}
\usepackage[hidelinks]{hyperref}
\usepackage[utf8]{inputenc}
\usepackage[small]{caption}
\usepackage{graphicx}
\usepackage{amsmath}
\usepackage{amsthm}
\usepackage{booktabs}
\usepackage{algorithm}
\usepackage{algorithmic}
\usepackage[switch]{lineno}
\usepackage{amssymb}

\usepackage{xcolor}
\usepackage{colortbl}
\usepackage{dblfloatfix}

\title{\textbf{TransSLR: A Lightweight Transformer for Sign Language Recognition}}

\author{
Wanchi Lucia Yen$^1$\and
Samuel Johnny$^1$\and
Victor Tolulope Olufemi$^1$\and
Emmanuel Aaron$^1$\and
Moise Busogi$^1$\\
\affiliations
$^1$Carnegie Mellon University Africa\\
\emails
\{wluciaye, sjohnny, volufemi, eaaron, mbusogi\}@andrew.cmu.edu
}

\begin{document}

\maketitle

\begin{abstract}
Automated sign language recognition for underrepresented languages remains a largely unsolved problem. Central African Sign Language (CASL) exemplifies this challenge: the CASL-W60 benchmark contains 60 isolated signs, and the best published result is 69.93\%. We present TransSLR, a lightweight Temporal Transformer Encoder trained from scratch on normalized 64-frame pose sequences. By operating on geometric keypoint representations rather than raw RGB, TransSLR achieves 80.39\% top-1 accuracy and 91.07\% top-5 accuracy on a signer-independent evaluation split. This result is 10.46 percentage points higher than the published 69.93\% result. TransSLR has 8.67 million trainable parameters. We also evaluate zero-shot transfer from a high-resource sign-language model, which achieves 0.00\% exact-match accuracy under our manual gloss-matching protocol. Our experiments compare pose-based, RGB-based, and multimodal approaches to sign-language recognition for a low-resource language.
\end{abstract}

\vspace{0.5cm} 

\section{Introduction}

Sign languages are rich and fully expressive linguistic systems, yet progress in 
Sign Language Processing (SLP) remains highly uneven across regions. While 
languages such as American Sign Language (ASL), British Sign Language (BSL), 
and Chinese Sign Language (CSL) benefit from large-scale datasets and 
well-established benchmarks, many low-resource sign languages remain severely 
underrepresented. Central African Sign Language (CASL), used across the Central African region, is one such example, where limited data availability restricts the development of robust recognition systems and hinders accessibility to 
AI-driven communication tools.

\begin{figure}[H]
    \centering
    \includegraphics[width=\linewidth]{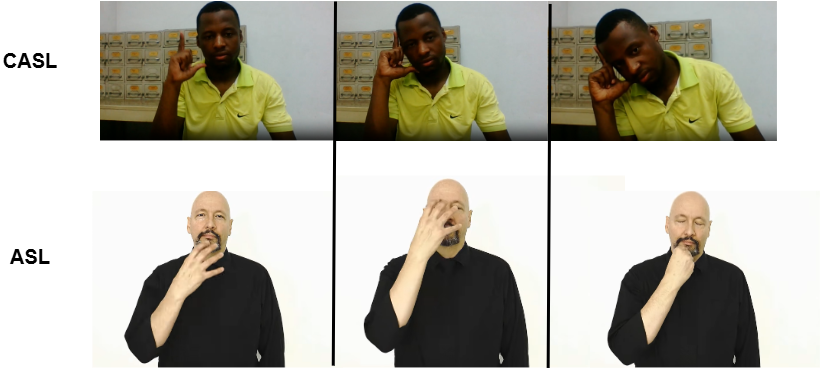}
    \caption{Example of lexical divergence between ASL and CASL. Although both signs convey ``sleep,'' they use distinct articulatory strategies, highlighting why knowledge learned from ASL does not transfer effectively to CASL.}
    \label{fig:lexical_divergence}
\end{figure}

Although ASL and CASL are both natural sign languages, they are not mutually interchangeable. They differ in vocabulary, articulation, and the linguistic strategies used to express the same concept. Figure~\ref{fig:lexical_divergence} illustrates one such difference using the sign for ``sleep". In ASL, the sign is produced with a downward hand-closing motion that iconically represents closing eyelids, whereas in CASL, the sign is expressed through a lateral head tilt that conveys the act of resting. This difference in sign formation illustrates linguistic divergence between the languages. The zero-shot result is evaluated separately using manual gloss matching.


Sign Language Recognition (SLR) is generally divided into isolated sign language recognition (ISLR) and continuous sign language recognition (CSLR) settings. Recent advances in ISLR have been driven by large-scale visual datasets and powerful architectures leveraging RGB video, multimodal inputs, and pre-training strategies. However, these approaches rely heavily on data abundance and often fail to generalize to low-resource settings with little corpus of data. In particular, the widely adopted assumption that models pre-trained on high-resource corpora (e.g., WLASL) can transfer effectively to unseen sign languages remains insufficiently validated.

In this work, we evaluate zero-shot transfer of sign-language recognition models to Central African Sign Language (CASL). Under the manual gloss mapping used in our experiment, the transferred model achieves 0.00\% exact-match accuracy. This observation is specific to that mapping and does not show that the pretrained model learned no transferable visual features.


To address this challenge, we propose TransSLR, a lightweight Transformer-based framework that emphasizes geometric representation learning over raw visual appearance. By operating on normalized skeletal pose sequences, 
our approach promotes invariance to signer-specific and environmental 
variations, making it particularly suitable for low-resource settings. Unlike RGB-dependent methods, TransSLR focuses on capturing the intrinsic motion dynamics of sign language.

    
    

We evaluate TransSLR on CASL-W60 using a signer-independent split. The model achieves 80.39\% accuracy on the 862-sample held-out signer group. This group was also used for checkpoint selection, so the score is not an untouched test estimate. The split prevents signer overlap between training and evaluation, addressing the leakage risk of random video-level splits.
Our main contributions are:
\begin{itemize}
    \item We identify and empirically characterize the \textit{overconfident 
    ignorance} phenomenon in cross-lingual transfer for sign language 
    recognition.
    
    \item We propose TransSLR, a Transformer-based ISLR framework 
    designed for low-resource settings, leveraging normalized skeletal 
    representations for improved generalization.
    
    \item We report 80.39\% accuracy on the signer-independent CASL-W60 evaluation set, 
    highlighting the advantages of geometric modeling over appearance-based 
    approaches in data-scarce environments.
\end{itemize}

\section{Related Work}

\subsection{Sign Language Recognition}

Sign Language Recognition (SLR) aims to interpret visual signing into linguistic representations and is commonly divided into isolated sign language recognition (ISLR) and continuous sign language recognition (CSLR). Early approaches relied on RGB video inputs, where convolutional neural networks (CNNs) such as I3D and S3D were used for spatial feature extraction, followed by recurrent neural networks (RNNs), including LSTMs and GRUs, for temporal modeling~\cite{hu2021signbert,hu2021handmodel,jiang2021skeletonmultimodal,vaezijoze2019msasl,zuo2022nla,10.24963/ijcai.2023/85}.

More recent work has shifted toward efficient and privacy-preserving representations using skeletal pose estimation. Frameworks such as MediaPipe~\cite{huynh2025motion} and OpenPose~\cite{8765346} enable the extraction of structured keypoints, which reduce redundancy while preserving motion dynamics. These representations have been effectively modeled using Graph Convolutional Networks (GCNs) to capture spatial dependencies~\cite{yang2026graph,jiang2021skeletonmultimodal}.

Transformer-based architectures have further improved temporal modeling by capturing long-range dependencies through self-attention~\cite{gan2023contrastive}. SPOTER~\cite{bohavcek2022sign} demonstrated the effectiveness of Transformers with spatial normalization, while more recent methods such as SNDA~\cite{hassan2025novel}, enhance attention mechanisms for sign recognition. Additionally, end-to-end approaches such as AutoSign~\cite{johnny2025autosign} explore direct pose-to-text translation for continuous sign language recognition, highlighting the growing importance of skeletal representations.

Large-scale datasets such as WLASL~\cite{li2020wlasl} and MS-ASL~\cite{vaezijoze2019msasl} have enabled the development of powerful spatiotemporal models, including VideoMAE~\cite{kumar2025videomae}. However, these datasets primarily focus on American Sign Language, limiting their applicability to other linguistic and regional contexts.

\subsection{Isolated Sign Language Recognition}

Isolated Sign Language Recognition (ISLR) focuses on classifying individual signs and serves as a fundamental component of broader SLR systems~\cite{sarhan2023unraveling,laines2023isolatedsignlanguagerecognition}. Traditional ISLR approaches combine CNN-based feature extraction with RNN-based temporal modeling, which can struggle with sequence variability and temporal alignment~\cite{10526274}.

Transformer-based methods have emerged as a strong alternative due to their flexibility in handling variable-length sequences. However, their effectiveness often depends on preprocessing strategies such as temporal sampling and spatial normalization.


For low-resource sign languages, the CASL-W60 dataset~\cite{lucky2025casl} provides the first benchmark for isolated Central African Sign Language recognition. Existing methods have demonstrated the feasibility of pose-based recognition on CASL, yet there remains limited understanding of how to effectively exploit kinematic representations for robust recognition in data-scarce settings. This work builds on these efforts by investigating whether a Transformer-based architecture operating on normalized kinematic pose sequences can improve recognition performance while maintaining a simple and efficient design.

\section{Methodology}
\subsection{Problem Formulation}
Given an input video sequence of a signer $\mathbf{F} = \{f_i\}_{i=1}^{N}$ consisting of $N$ frames, the goal of isolated sign language recognition (ISLR) is to learn a mapping function $\mathcal{G}$ that predicts the corresponding sign label $y \in \mathcal{Y}$, where $\mathcal{Y}$ denotes the predefined vocabulary of sign classes.

Formally, the task can be expressed as:
\[
y = \mathcal{G}(\mathbf{F})
\]

where $\mathcal{G}$ represents the end-to-end model that encodes spatio-temporal information from the input video and outputs the predicted sign.

\subsection{Data Preprocessing}

To obtain a compact kinematic representation, each video is
processed using MediaPipe Holistic. For every selected frame,
we retain 33 body-pose landmarks together with 21 landmarks
from each hand, yielding 75 landmarks with $(x,y,z)$ coordinates
and therefore a 225-dimensional frame-level representation.

Videos are standardized to a maximum sequence length of
$T=64$ frames using fixed-stride temporal sampling. Missing
landmark detections are initially represented as zeros. After
constructing the 64-frame sequence, missing entries are replaced
using forward filling followed by backward filling, allowing
temporally adjacent valid detections to preserve continuity in the
kinematic trajectory.

Before being passed to the model, each sequence is spatially
normalized. Let $P_{t,i}\in\mathbb{R}^{3}$ denote landmark $i$
at frame $t$. We first center all landmarks relative to the
MediaPipe nose landmark (pose landmark 0):

\begin{equation}
\tilde{P}_{t,i} = P_{t,i} - P_{t,0}.
\end{equation}

To reduce variation due to signer scale, we compute a
sequence-level scale factor using the mean Euclidean distance
between the left and right shoulder landmarks (pose landmarks
11 and 12):

\begin{equation}
s = \frac{1}{T}\sum_{t=1}^{T}
\left\|
\tilde{P}_{t,11} - \tilde{P}_{t,12}
\right\|_2 .
\end{equation}

The final normalized landmarks are then given by

\begin{equation}
X_{t,i} = \frac{\tilde{P}_{t,i}}{s},
\end{equation}
resulting in a normalized input tensor
$X\in\mathbb{R}^{64\times225}$ for each video.
\begin{figure}[htbp]
    \centering
    \includegraphics[width=0.9\linewidth]{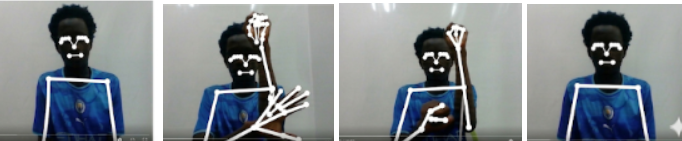}
    \caption{The RGB and 64-Frame Motion Ribbon. A horizontal time-strip showing the centralized progression of a sign from onset to offset.}
    \label{fig:tradslr}
\end{figure}

\subsection{TransSLR Architecture}

\subsubsection{Overview}

TransSLR is a lightweight Transformer-based encoder designed
for Isolated Sign Language Recognition (ISLR) under low-resource
constraints. Given a normalized pose sequence, the model learns
relationships among frame-level kinematic representations and
produces a prediction over the 60 CASL-W60 sign classes.

The architecture consists of three main stages: (1) frame-level
feature projection, (2) a stack of Transformer encoder layers,
and (3) temporal aggregation followed by a classification head.
Unlike sequence-to-sequence tasks such as Continuous Sign
Language Recognition (CSLR) or Sign Language Translation (SLT),
ISLR requires a single class prediction for each bounded sign
sequence. Therefore, TransSLR uses an encoder-only architecture
without an autoregressive decoder.

\subsubsection{Input Representation}

Each input sample is represented as a kinematic tensor

\begin{equation}
    X \in \mathbb{R}^{T \times D},
\end{equation}

where $T=64$ denotes the number of frames and $D=225$
corresponds to 75 skeletal landmarks, each represented by
three-dimensional $(x,y,z)$ coordinates.

As described in Section~3.2, the pose coordinates are centered
relative to the nose landmark and scaled using the mean
shoulder-to-shoulder distance across the sequence. This
normalization reduces positional and body-scale variation while
preserving the geometric structure of the signing motion.

\subsubsection{Frame-Level Feature Projection}

The 225-dimensional pose vector at each frame is projected into
a 512-dimensional latent representation using a learned linear
transformation. For an input frame $x_t \in \mathbb{R}^{225}$,
the projection is defined as

\begin{equation}
    h_t = W_p x_t + b_p,
\end{equation}

where $W_p \in \mathbb{R}^{512 \times 225}$ and $b_p$ denote
the learned projection parameters.

The projected sequence is processed by BatchNorm1d with 64 channels, followed by ReLU activation and dropout with probability $p=0.3$. This produces the frame-level representation

\begin{equation}
    H^{(0)} \in \mathbb{R}^{64 \times 512},
\end{equation}

which is passed to the Transformer encoder.

\subsubsection{Transformer Encoder Stack}

The projected pose sequence is processed by four stacked
Transformer encoder layers. Each encoder layer has a model
dimension of $d_{\mathrm{model}}=512$, eight self-attention heads,
and a feed-forward hidden dimension of 1024.

The final 80.39\% model does not add learned or sinusoidal positional embeddings to the Transformer input. The frame-level BatchNorm1d layer has 64 channels. 

For each encoder layer, Multi-Head Self-Attention (MHSA)
allows each frame representation to interact with representations
from the other frames in the sequence. The attention output is
followed by a position-wise feed-forward network (FFN), with
residual connections, layer normalization, and dropout applied
within the encoder block.

The encoder transformation can be summarized as

\begin{equation}
    H^{(\ell)}
    =
    \mathrm{TransformerEncoder}_{\ell}
    \left(H^{(\ell-1)}\right),
    \qquad \ell = 1,\ldots,4.
\end{equation}

Dropout with probability $p=0.3$ is applied within the
Transformer layers. The final encoder output is therefore

\begin{equation}
    H^{(4)} \in \mathbb{R}^{64 \times 512}.
\end{equation}

This encoder-only formulation provides a compact mechanism for
learning interactions among the kinematic representations of the
64-frame sign sequence without the additional complexity of a
decoder.

\subsubsection{Temporal Aggregation and Classification Head}

Following the Transformer encoder, Global Average Pooling (GAP)
aggregates the frame-level representations into a single
sequence-level embedding:

\begin{equation}
    h_{\mathrm{GAP}}
    =
    \frac{1}{T}
    \sum_{t=1}^{T} H^{(4)}_t,
\end{equation}

where $h_{\mathrm{GAP}} \in \mathbb{R}^{512}$.

The pooled representation is passed through a two-layer
classification head. The first linear layer maps the
512-dimensional embedding to a 256-dimensional hidden
representation, followed by a ReLU activation and dropout with
$p=0.3$. A final linear layer maps the resulting representation
to logits over the 60 sign classes:

\begin{equation}
    z =
    W_2
    \left[
    \mathrm{Dropout}
    \left(
    \mathrm{ReLU}
    \left(
    W_1 h_{\mathrm{GAP}} + b_1
    \right)
    \right)
    \right]
    + b_2,
\end{equation}

where $z \in \mathbb{R}^{60}$. During inference, the logits are
converted to class probabilities using the Softmax function,

\begin{equation}
    \hat{y}_c =
    \frac{\exp(z_c)}
    {\sum_{j=1}^{60}\exp(z_j)},
\end{equation}

and the class with the highest probability is selected as the
predicted sign.

\subsubsection{Training Objective}

TransSLR is trained using class-weighted cross-entropy with label smoothing $\epsilon=0.1$. For a batch of $B$ examples, let $p_{i,c}$ denote the predicted probability for class $c$, $w_c$ its training-derived class weight, and $y_i$ the target class. The smoothed target is $\tilde y_{i,c}=(1-\epsilon)\mathbb{1}[c=y_i]+\epsilon/60$. With the weighted mean reduction used by PyTorch CrossEntropyLoss, the objective is
\begin{equation}
    \mathcal{L}_{\mathrm{WCE}} = \frac{1}{\sum_{i=1}^{B}w_{y_i}}\sum_{i=1}^{B}\left[-\sum_{c=1}^{60}w_c\tilde y_{i,c}\log(p_{i,c})\right].
\end{equation}
The class weights are computed from the training data.
\begin{figure*}[htbp]
    \centering
    \includegraphics[width=\textwidth]{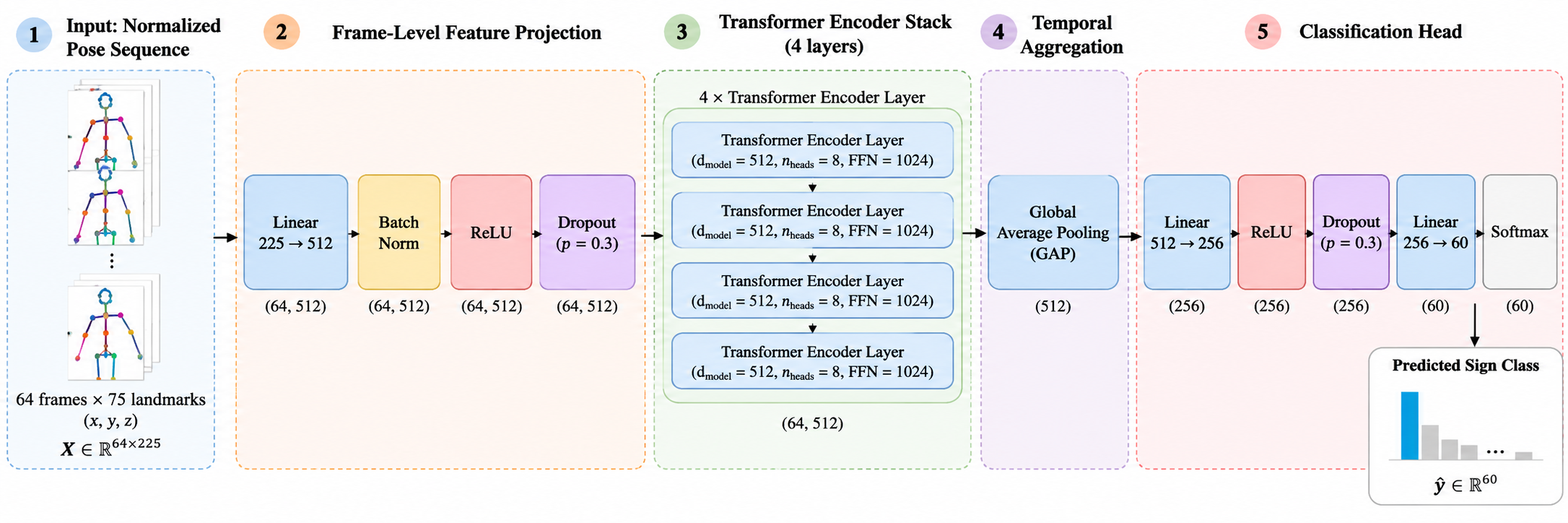}
    \caption{TransSLR architecture. A normalized 64-frame pose sequence with 225-dimensional frame-level features is projected to a 512-dimensional representation using a linear layer, BatchNorm1d, ReLU activation, and dropout. The resulting sequence is processed by four Transformer encoder layers with $d_{\mathrm{model}}=512$, eight attention heads, and a feed-forward dimension of 1024. Global average pooling aggregates the temporal representations into a 512-dimensional sequence embedding, which is passed through a two-layer classification head ($512 \rightarrow 256 \rightarrow 60$). The final logits are converted to class probabilities using Softmax during inference to predict one of the 60 CASL-W60 sign classes.}
    \label{fig:transnet_pipeline}
\end{figure*}

\section{Experimental Setup}

\subsection{Dataset}
Our experiments use CASL-W60, a benchmark of 60 isolated signs. The published dataset contains 5,889 samples from 19 signers and its original folder split has 3,667 training, 679 validation, and 1,543 test samples. This video-level random split permits signer overlap across partitions. The extracted feature files used for the signer-independent TransSLR run contain 14 distinct signer IDs.

For the reported TransSLR run, we applied GroupShuffleSplit with test size 0.2 and random state 42 to the 5,889 extracted files, grouping by signer. This yielded 5,027 training examples from 11 signers and 862 held-out examples from S01, S10, and S12, with no signer overlap. There was no separate validation partition: the 862-example signer-independent set was evaluated during training and used for checkpoint selection. Accordingly, 80.39\% is a model-selection score on a signer-disjoint set, not an untouched test estimate.

\subsection{Implementation Details}
The 80.39\% TransSLR run was trained in PyTorch on a single NVIDIA A100 GPU for up to 60 epochs with batch size 64. Inputs contain 64 frames and 225 coordinates per frame. We used AdamW with initial learning rate $1\times10^{-4}$ and weight decay 0.05, with a OneCycle learning-rate schedule peaking at $5\times10^{-4}$. Training used class-weighted cross-entropy with label smoothing 0.1, automatic mixed precision, and gradient clipping at norm 1.0. The checkpoint was selected using accuracy on the signer-independent 862-example set.

\subsection{Evaluation Protocol}
We report top-1 accuracy, top-5 accuracy, and top-1 error. For closed-set classification, top-1 error is the complement of top-1 accuracy:
\begin{equation}
    \text{Top-1 error} = 1 - \text{Acc}_{\text{top-1}}.
\end{equation}
This metric is not character error rate (CER) or word error rate (WER). The CASL-W60 source reports its single-sign accuracy as $1-\mathrm{WER}$; we retain its published score as reported. The signer-independent evaluation set for TransSLR was used for checkpoint selection, so its score is not an untouched test estimate.

\section{Results}

\begin{table*}[!t]
\centering
\setlength{\tabcolsep}{8pt}
\renewcommand{\arraystretch}{1.15}
\small
\begin{tabular}{lcccc}
\toprule
\textbf{Method} & \textbf{Modality} & \textbf{Top-1 (\%)} $\uparrow$ & \textbf{Top-5 (\%)} $\uparrow$ & \textbf{Top-1 error (\%)} $\downarrow$ \\
\midrule
VideoMAE~\cite{kumar2025videomae} (Zero-Shot)  & RGB      & 0.00  & -- & 100.00 \\
VideoMAE~\cite{kumar2025videomae} (Fine-Tuned) & RGB      & 57.55 & -- & 42.45  \\
\midrule
SLT~\cite{lucky2025casl}            & Pose     & 69.93 & -- & 30.07  \\
\midrule
Bi-GRU + Attention                             & Pose     & 70.34 & -- & 29.66  \\
Transformer (with MHSA + Positional Encoding)  & Pose     & 70.24 & -- & 29.76  \\
Enhanced Transformer (temporal + pooling)      & Pose     & 73.14 & -- & 26.86  \\
Transformer + MLP Head                         & Pose     & 75.82 & -- & 24.18  \\
Transformer Encoder (Final Baseline)           & Pose     & \textbf{76.13} & -- & \textbf{23.87} \\
MViT V2 + Bi-LSTM                              & RGB+Pose & 74.23 & -- & 25.77  \\
\midrule
\textbf{TransSLR (Ours)}                       & Pose     & \textbf{80.39} & \textbf{91.07} & \textbf{19.61} \\
\bottomrule
\end{tabular}

\caption{Results on CASL-W60. Top-1 denotes standard classification accuracy, while Top-5 measures whether the correct label appears within the top five predictions. $\uparrow$ indicates higher is better and $\downarrow$ lower is better. Top-5 is not reported for prior work.}

\label{tab:results}
\vspace{-3mm}
\end{table*}

\subsection{Fine-Tuning Evaluation and Domain Gap}
Table~\ref{tab:results} presents the zero-shot and fine-tuned performance of VideoMAE on CASL-W60. Under zero-shot evaluation, VideoMAE obtains 0.00\% exact-match accuracy under the manual gloss mapping used in this evaluation; this result does not establish that the pretrained representation contains no transferable visual information. 

Fine-tuning on the CASL-W60 training split recovers performance to
57.55\%, yet this remains approximately 12 points below the published CASL-W60 result~\cite{lucky2025casl} and over 22 points below
our proposed TransSLR. We attribute this persistent gap to a representational mismatch: VideoMAE encodes spatiotemporal appearance
features that do not transfer to CASL-W60, where lexical distinctions are carried primarily by subtle kinematic differences, handshape transitions, movement velocity, and spatial trajectory rather than by signer appearance. At the scale of CASL-W60, the available target-domain signal is insufficient to re-orient the model's inductive priors toward the kinematic structure that isolated sign discrimination requires.

\subsection{Comparison with Previously Reported Results}

TransSLR obtains 80.39\% accuracy (19.61\% top-1 error) on the signer-independent evaluation split for CASL-W60 (862 samples), which was also used for checkpoint selection (Table~\ref{tab:results}). This is a numerical difference of 10.46 percentage points from the published 69.93\% result and 4.26 percentage points from the 76.13\% video-level baseline. The 69.93\% and 76.13\% results use different evaluation protocols, so these differences are numerical and not direct comparisons.

TransSLR also achieves 91.07\% top-5 accuracy. Overall, the results demonstrate that the proposed encoder-only pose-based transformer provides a strong trade-off between accuracy and computational efficiency compared to more complex multimodal or RGB-heavy architectures.

\subsection{Retrieval Results Across Modalities}

\begin{figure}[htbp]
    \centering
    \includegraphics[width=0.5\textwidth]{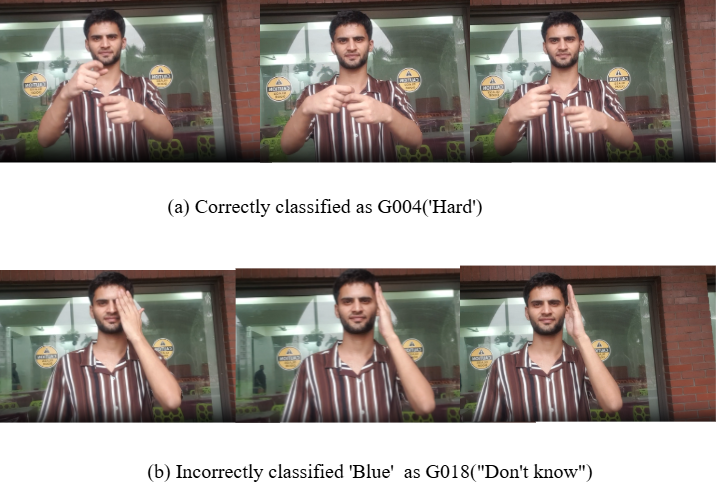}
    \caption{ Qualitative results of the \textbf{TransSLR} model}
    \label{fig:qualr}
\end{figure}

Table~\ref{tab:results} reports closed-set classification performance under the evaluation protocols indicated for each method. Table~\ref{tab:robustness_results} reports retrieval metrics from the signer-independent retrieval evaluation. The MViT V2 + Bi-LSTM obtains Recall@1 of 29.90\% in that evaluation. TransSLR's retrieval scores are also reported in Table~\ref{tab:robustness_results}. These retrieval results and Table~\ref{tab:results}'s classification accuracies are different metrics and should not be compared as if they were the same evaluation.

The observed difference does not establish why the models perform differently. Possible factors include input modality, architecture, and training data, but the experiments do not isolate these effects or test whether the RGB model relies on signer-specific appearance cues.

TransSLR uses nose-centering and shoulder-distance scaling to reduce positional and body-scale variation in pose inputs. The retrieval evaluation reports Recall@10 of 93.62\% and Median Rank of 1.0 for TransSLR, compared with Median Rank of 4.0 for the multimodal baseline. These findings describe the reported retrieval experiment and do not establish superiority across evaluation settings.
\subsection{Failure Analysis}

Despite high global accuracy, a granular evaluation of the precision-recall matrix reveals localized failure patterns. Specifically, five classes (30, 35, 51, 57, and 59) returned a 0.00\% F1-score. Rather than indicating random noise, this pattern highlights classes for which the model fails to recover the correct decision boundaries. These errors are further reflected in the corresponding false-positive distributions. For example, Class 53 achieved a perfect recall of 1.0000 but a severely degraded precision of 0.3333. This indicates an over-prediction pattern, where multiple samples are assigned to Class 53 despite belonging to other classes. Such errors may arise when visually similar signs exhibit overlapping pose trajectories or when discriminative information is not sufficiently represented by the pose-only input. In these cases, the spatial encoder may rely on shared motion patterns rather than features that distinguish individual signs. These findings suggest that while skeletal normalization reduces signer-positioning variation, future iterations could investigate facial and other non-manual features to determine whether they improve recognition of overlapping signs.

\begin{table}[t]
\centering
\footnotesize
\setlength{\tabcolsep}{3pt}
\renewcommand{\arraystretch}{1.1}

\begin{tabular}{l l c c c c}
\hline
\textbf{Mod.} & \textbf{Model} & \textbf{R@1} & \textbf{R@5} & \textbf{R@10} & \textbf{MedR} \\
\hline
Mult. & MViT-BiLSTM & 29.90\% & 57.00\% & 73.78\% & 4.0 \\
Pose & \textbf{TransSLR} & \textbf{80.39\%} & \textbf{91.07\%} & \textbf{93.62\%} & \textbf{1.0} \\
\hline
\end{tabular}
\caption{Evaluation of retrieval performance on unseen signers using Recall@K and Median Rank. Results show that TransSLR consistently has higher reported retrieval scores than the multimodal baseline in this experiment.}
\label{tab:robustness_results}
\end{table}

\subsection{Qualitative Analysis of TransSLR}

Figure~\ref{fig:qualr} illustrates selected correct and incorrect predictions from TransSLR on the signer-independent evaluation set, which was used for checkpoint selection. In Figure~\ref{fig:qualr}(a), 
the model correctly classifies the sign (\textit{Hard}), demonstrating 
its ability to capture discriminative hand shape and motion trajectories from 
normalized pose sequences. Figure~\ref{fig:qualr}(b) shows a failure 
case where \textit{Blue} is misclassified as (\textit{Don't know}), 
suggesting that visually similar kinematic trajectories between certain sign 
pairs remain a challenge under the limited training data available in CASL-W60. 
This highlights the need for expanded vocabulary annotation as a priority for 
future data curation efforts.

\subsection{Latent Manifold Integrity and Signer Invariance}

\begin{figure}[!htbp]
    \centering
    \includegraphics[width=1.0\linewidth]{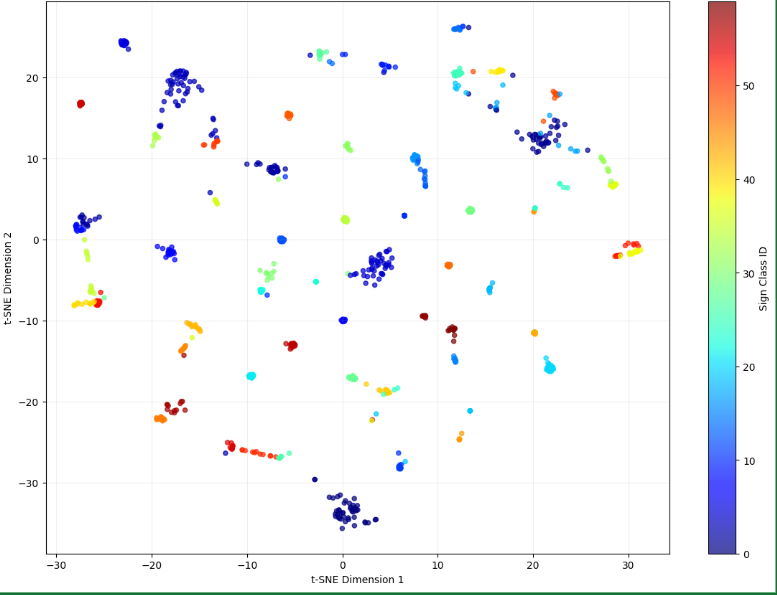}
    \caption{t-Distributed Stochastic Neighbor Embedding (t-SNE) visualization of the TransSLR latent manifold. The plot represents 862 signer-disjoint evaluation samples projected from the penultimate layer. This set was used for checkpoint selection. The visualization is qualitative and shows apparent class structure with overlapping regions; it does not establish signer invariance or semantic separability.}
    \label{fig:tsne_clusters}
\end{figure}

The t-SNE projection of the TransSLR latent representations (Figure~\ref{fig:tsne_clusters}) provides a qualitative view of the learned embeddings. Apparent clusters and overlaps depend on the projection; the figure does not independently establish class separability or signer invariance.

We observe failure modes in a small number of classes that are visually similar in pose trajectory but differ in non-manual components such as facial expressions. Since the current model operates on pose-only inputs, these results suggest that incorporating facial or non-manual features may improve discrimination in these cases.

\begin{table}[htbp]
\centering
\footnotesize
\setlength{\tabcolsep}{3pt}
\renewcommand{\arraystretch}{1.10}
\begin{tabular}{lcc}
\toprule
\textbf{Method} & \textbf{Params (M)} $\downarrow$ & \textbf{FLOPs (G)} $\downarrow$ \\
\midrule
VideoMAE (Fine-Tuned)             & 65.021  & 101.85 \\
Bi-GRU + Attention              & 9.95  & 1.40 \\
MViT V2 + Bi-LSTM           & 47.47  &  7.17\\
\midrule
\textbf{TransSLR (Ours)}    & \textbf{8.67} & \textbf{0.28} \\
\bottomrule
\end{tabular}%
\caption{Comparison of computational complexity across evaluated models. Parameter count is computed over all trainable parameters, while FLOPs are measured for processing a single 64-frame pose sequence. $\downarrow$ indicates lower is better.}
\label{tab:efficiency}
\vspace{-3mm}
\end{table}

\subsection{Efficiency Analysis}

\noindent We evaluate the computational efficiency of TransSLR by comparing its model size and inference complexity against representative RGB-based, pose-based, and multimodal baselines. As shown in Table~\ref{tab:efficiency}, TransSLR has the lowest parameter count among the listed models, with 8.67M trainable parameters, while also exhibiting the lowest computational cost at 0.277 GFLOPs per 64-frame pose sequence. In comparison, VideoMAE and MViT V2 + Bi-LSTM require substantially larger models (65.02M and 47.47M parameters) and higher computational cost (101.85 and 7.17 GFLOPs, respectively). These reported estimates suggest a compact model. The retrieval comparisons and efficiency estimates should be interpreted according to their respective evaluation and measurement protocols.

\section*{Conclusion and Future Work}
We presented TransSLR, a lightweight Transformer encoder for isolated sign language recognition using normalized pose sequences. On the signer-independent CASL-W60 evaluation set, it achieved 80.39\% top-1 accuracy and 91.07\% top-5 accuracy. Signers in this set were excluded from training, but the set was also used to select the checkpoint; these results therefore measure performance on a signer-disjoint model-selection set, not an untouched test set. The 80.39\% score is numerically 10.46 percentage points higher than the published 69.93\% result, although the evaluation protocols differ.

The model uses nose-centering and shoulder-distance scaling to reduce positional and body-scale variation in the input. The current results support further study of pose-based recognition for low-resource sign languages, while a final evaluation with separate signer-disjoint validation and test sets is needed to estimate generalization without checkpoint-selection bias.

Future work includes expanding annotated CASL corpora and extending recognition from isolated signs to continuous signing. Continuous recognition will require modeling unsegmented streams, long-range temporal dependencies, and variable-length outputs, potentially using Connectionist Temporal Classification (CTC) or attention-based sequence decoding. Self-supervised learning and larger-scale pretraining may also improve data efficiency and support work on other low-resource African sign languages.

\bibliographystyle{named}
\bibliography{ijcai26}

\end{document}